\documentclass[10pt,conference]{IEEEtran}

\usepackage{amsmath, amssymb}
\usepackage{graphicx}
\usepackage{booktabs}
\usepackage{multirow}
\usepackage{array}
\usepackage{xcolor}
\usepackage[hidelinks,hypertexnames=false]{hyperref}
\usepackage{url}
\usepackage{enumitem}
\usepackage{balance}
\usepackage{algorithm}
\usepackage{algpseudocode}
\usepackage{placeins}
\usepackage{orcidlink}

\hypersetup{
  pdftitle={Moment-Guided Optimizers for Noisy Deep Learning: MGO-SC as a Simple Consistency Gate for Adam-Style Updates},
  pdfauthor={Alicem Koyun},
  pdfsubject={Accepted at the 2026 IEEE Global Emerging Technologies Conference (iGET)},
  pdfkeywords={Optimization, adaptive methods, noisy labels, robustness, Adam, ResNet, CIFAR-10}
}

\makeatletter
\def\ps@IEEEtitlepagestyle{%
  \def\@oddhead{}\def\@evenhead{}%
  \def\@oddfoot{\parbox[t]{\textwidth}{\scriptsize Accepted for publication at the 2026 IEEE Global Emerging Technologies Conference (iGET), Irvine, CA, USA.\\
\copyright~2026 IEEE. Personal use of this material is permitted. Permission from IEEE must be obtained for all other uses, in any current or future media, including reprinting/republishing this material for advertising or promotional purposes, creating new collective works, for resale or redistribution to servers or lists, or reuse of any copyrighted component of this work in other works.}}%
  \def\@evenfoot{}%
}
\makeatother

\newcommand{\maybeincludegraphics}[2][]{%
  \IfFileExists{#2}{\includegraphics[#1]{#2}}{%
    \fbox{\parbox{0.95\linewidth}{\centering\small Figure file unavailable: \texttt{#2}}}%
  }%
}

\title{Moment-Guided Optimizers for Noisy Deep Learning: MGO-SC as a Simple Consistency Gate for Adam-Style Updates}

\author{
  \IEEEauthorblockN{Alicem Koyun\,\orcidlink{0009-0005-5152-3500}}
  \IEEEauthorblockA{Master of Science in Computer Science\\
  Westcliff University, Irvine, USA\\
  Email: alicemkoyun@gmail.com}
}

\begin{document}
\maketitle

\begin{abstract}
Adaptive optimizers are widely used in deep learning, yet their behavior can degrade under heavy stochasticity and label noise where the sign and magnitude of gradients can be unstable. We study \textbf{MGO-SC} (Moment-Guided Optimizer with \emph{Sign-Consistency}), a lightweight modification of Adam-style methods that derives a coordinate-wise \emph{consistency gate} from the magnitude of a sign exponential moving average (EMA): coordinates whose gradient sign is stable are trusted, while coordinates whose sign fluctuates are down-weighted. We show that the naive form of this gate, which can only \emph{shrink} updates, loses effective step size and \emph{underperforms} when the learning rate is matched to the baseline. We therefore introduce a \emph{mean-preserving} variant that reallocates trust across coordinates without changing the net per-tensor step magnitude. On \textbf{CIFAR-10} with \textbf{ResNet-18}, under a matched learning rate and decoupled weight decay, mean-preserving MGO-SC \emph{consistently improves over its AdamW baseline} across clean, symmetric, and asymmetric label noise (e.g., $+1.4$ points at $40\%$ symmetric noise; paired bootstrap $95\%$ CI excluding zero; Wilcoxon $p<0.01$), while adding no second-order information or per-sample filtering. We report matched-seed effect sizes, nonparametric bootstrap confidence intervals, and paired significance tests, and we release code and raw logs.
\end{abstract}


\section{Introduction}
Modern deep learning pipelines frequently default to adaptive optimizers such as Adam \cite{kingma2015adam} and AdamW \cite{loshchilov2019adamw} due to their stability and fast convergence. However, ``stability'' is not guaranteed under all forms of stochasticity. In particular, \emph{label noise} changes the gradient distribution by injecting contradictory supervision signals; this increases gradient variance and can raise the probability of coordinate-wise sign flips. In practice, such instability may cause effective stepsizes to become overly aggressive in directions where the optimizer's moment statistics are corrupted by inconsistent gradient signs.

This work focuses on a narrow but common regime: supervised learning with symmetric label noise (random label flips). Rather than proposing a new loss, sample selection strategy, or multi-network filtering pipeline (common in noisy-label learning \cite{song2023noisylabels_survey}), we ask a simpler question: \emph{Can we add a minimal mechanism inside an Adam-style optimizer to attenuate updates when the gradient direction is unreliable?}

\textbf{Contribution.} We study \textbf{MGO-SC}, which adds a \emph{sign-consistency gate} to Adam-style updates, computed from a sign exponential moving average (EMA) of the gradient. Our central finding is that \emph{how} the gate is applied is decisive: a naive gate that only shrinks updates loses effective step size and, under a matched learning rate, underperforms the AdamW baseline; a \emph{mean-preserving} gate that reallocates trust across coordinates while keeping the average step magnitude fixed instead yields a small but consistent and statistically significant improvement over AdamW. The gate uses only elementwise operations on tensors already maintained by Adam and requires no additional forward/backward passes.

\textbf{Empirical evidence.} We provide controlled experiments on CIFAR-10 \cite{krizhevsky2009cifar} using ResNet-18 \cite{he2016resnet}. We report mean\,$\pm$\,std over multiple seeds and include paired-seed comparisons and bootstrap intervals to avoid over-interpreting small numerical gaps.

\textbf{Paper organization.} Section~\ref{sec:related} reviews adaptive optimizers and noisy-label learning. Section~\ref{sec:method} describes MGO-SC and its rationale. Section~\ref{sec:experiments} details the experimental protocol and results. Section~\ref{sec:discussion} discusses diagnostics, limitations, and directions for stronger validation.

\textbf{Open science.} Code and raw logs are available at \url{https://github.com/alicemkyn/moment_guided_optimizers_pytorch}. This includes the training code, configuration files, and raw CSV logs needed to reproduce the tables and figures in this paper.

\section{Related Work}\label{sec:related}

\subsection{Adaptive optimizers and modern variants}
Adam \cite{kingma2015adam} introduced exponentially-weighted first and second moments with bias correction. Subsequent work improved theoretical properties (e.g., AMSGrad \cite{reddi2018adam}) and practical generalization (e.g., decoupled weight decay in AdamW \cite{loshchilov2019adamw}). Recent proposals often aim to reduce tuning burden, improve stability, or scale to modern training regimes. Examples include AdaBelief \cite{zhuang2020adabelief}, which adapts stepsizes based on a "belief" view of gradient deviations, LAMB \cite{you2019lamb} for large-batch training, and Adan \cite{xie2022adan} which reformulates Nesterov-style momentum within an adaptive framework. Other directions include variance rectification (RAdam \cite{liu2019radam}), lookahead-style meta-updates (Lookahead \cite{zhang2019lookahead}), curvature-aware preconditioning (AdaHessian \cite{yao2020adahessian}), and optimization-centric regularizers such as gradient centralization \cite{yong2020gc}. There is also growing interest in simplified sign-based updates such as Lion \cite{chen2023lion} and scalable second-order approaches such as Sophia \cite{liu2023sophia}. Finally, learning-rate automation is explored by D-Adaptation \cite{defazio2023dadaptation}.

MGO-SC is intentionally lightweight relative to these optimizers: we keep the Adam-style moment structure and add a single gate term designed specifically to react to sign instability that may arise under label noise.

\subsection{Generalization and sharpness-aware training}
Optimizer choice can affect generalization through implicit regularization and loss landscape geometry. Sharpness-Aware Minimization (SAM) \cite{foret2021sam} explicitly encourages flat minima and often improves robustness, at the cost of extra forward/backward passes per step. While orthogonal to our goal, SAM motivates the broader theme that small modifications to the update rule can yield meaningful robustness gains. In contrast, MGO-SC targets a different failure mode (sign instability under noisy supervision) without additional compute.

\subsection{Learning with noisy labels}
Learning from noisy labels has been studied extensively; surveys by Song \emph{et al.} \cite{song2023noisylabels_survey} and Algan and Ulusoy \cite{algan2021noisylabels_survey} summarize core method families and evaluation pitfalls. Representative families include robust losses such as generalized cross entropy (GCE) \cite{zhang2018gce} and symmetric cross entropy (SCE) \cite{wang2019sce}, curriculum and peer-learning methods such as Co-teaching \cite{han2018coteaching}, and semi-supervised mixing pipelines such as DivideMix \cite{li2020dividemix}. Early-learning regularization (ELR) \cite{liu2020elr} reduces memorization of corrupted labels by discouraging confident predictions early in training. Additional strong baselines include variance-regularized training \cite{luo2019noisylabels_baseline} and augmentation-centered strategies \cite{nishi2021augmentation_lnl}. Confident learning \cite{northcutt2021confidentlearning} provides principled tools to estimate label issues and supports practical data cleaning.

MGO-SC is not a replacement for these approaches. Instead, it provides an optimizer-side mechanism that can be combined with robust losses or filtering methods. Our emphasis is on a minimal change that is easy to reproduce and integrate.

\section{Method: MGO-SC}\label{sec:method}

\subsection{From Adam to sign-consistency gating}
Let $g_t = \nabla_{\theta} \mathcal{L}(\theta_t)$ be the stochastic gradient at step $t$.
Adam maintains exponential moving averages (EMAs):
\begin{align}
  m_t &= \beta_1 m_{t-1} + (1-\beta_1) g_t,\\
  v_t &= \beta_2 v_{t-1} + (1-\beta_2) g_t^2.
\end{align}
With bias correction $\hat m_t = m_t / (1-\beta_1^t)$ and $\hat v_t = v_t / (1-\beta_2^t)$, Adam updates
\begin{equation}
  \theta_{t+1} = \theta_t - \alpha \frac{\hat m_t}{\sqrt{\hat v_t}+\epsilon}.
  \label{eq:adam}
\end{equation}

Under label noise, the gradient direction can become inconsistent across minibatches. MGO-SC maintains a \emph{sign EMA}
\begin{equation}
  s_t = \beta_s\, s_{t-1} + (1-\beta_s)\,\mathrm{sign}(g_t),
  \label{eq:signema}
\end{equation}
whose magnitude $|\hat s_t|$ (optionally bias-corrected) measures how consistently each coordinate's gradient sign points in one direction: $|\hat s_t|\to 1$ for sign-stable coordinates and $|\hat s_t|\to 0$ when signs flip frequently. From this we form a coordinate-wise reliability
\begin{equation}
  r_t = \max\!\big(\,|\hat s_t|^{\gamma},\; r_{\min}\big),
  \label{eq:reliability}
\end{equation}
with exponent $\gamma\ge 0$ and floor $r_{\min}\in[0,1)$. The applied gate $s^{\mathrm{gate}}_t$ is obtained from $r_t$ in one of two ways,
\begin{align}
  \text{(shrink)}\quad & s^{\mathrm{gate}}_t = \min(r_t,\,1), \label{eq:gate-shrink}\\
  \text{(mean-preserving)}\quad & s^{\mathrm{gate}}_t = r_t \,/\, \overline{r_t}, \label{eq:gate-renorm}
\end{align}
where $\overline{r_t}$ is the mean of $r_t$ over the coordinates of a parameter tensor, and the step is
\begin{equation}
  \theta_{t+1} = \theta_t - \alpha \, s^{\mathrm{gate}}_t \odot \frac{\hat m_t}{\sqrt{\hat v_t}+\epsilon}.
  \label{eq:mgosc}
\end{equation}
The shrink gate \eqref{eq:gate-shrink} can only reduce step magnitude, so it lowers the \emph{effective} learning rate whenever many coordinates are down-weighted. The mean-preserving gate \eqref{eq:gate-renorm} keeps the average per-tensor multiplier equal to one, \emph{redistributing} trust from sign-unstable to sign-stable coordinates without changing the net step size. Section~\ref{sec:experiments} shows this distinction determines whether the gate helps or hurts. A binary variant that replaces $|\hat s_t|$ with the instantaneous agreement indicator $\mathbb{I}[\mathrm{sign}(g_t)=\mathrm{sign}(m_t)]$ behaves like the shrink gate and is reported as an ablation. When $\gamma=0$, $r_t\equiv 1$ and MGO-SC reduces to the base optimizer.

\begin{algorithm}[t]
\caption{MGO-SC (Moment-Guided Optimizer with Sign-Consistency)}
\label{alg:mgosc}
\begin{algorithmic}[1]
\Require Parameters $\theta_0$, stepsize $\alpha$, betas $(\beta_1,\beta_2)$, sign smoothing $\beta_s$, exponent $\gamma$, floor $r_{\min}$, mode $\in\{$shrink, mean-preserving$\}$, $\epsilon$
\State Initialize $m_0 \gets 0$, $v_0 \gets 0$, $s_0 \gets 0$
\For{$t=1,2,\dots$}
  \State $g_t \gets \nabla_{\theta}\mathcal{L}(\theta_{t-1})$
  \State $m_t \gets \beta_1 m_{t-1} + (1-\beta_1)g_t$
  \State $v_t \gets \beta_2 v_{t-1} + (1-\beta_2)(g_t \odot g_t)$
  \State $\hat m_t \gets m_t/(1-\beta_1^t)$,\quad $\hat v_t \gets v_t/(1-\beta_2^t)$
  \State $s_t \gets \beta_s s_{t-1} + (1-\beta_s)\,\mathrm{sign}(g_t)$;\quad $\hat s_t \gets s_t$ (opt.\ bias-corrected)
  \If{$\gamma = 0$}
    \State $s^{\mathrm{gate}}_t \gets \mathbf{1}$ \Comment{reduces to base update}
  \Else
    \State $r_t \gets \max(|\hat s_t|^\gamma,\, r_{\min})$
    \State $s^{\mathrm{gate}}_t \gets \min(r_t,1)$ \textbf{if} shrink \textbf{else} $r_t/\overline{r_t}$
  \EndIf
  \State $\theta_t \gets \theta_{t-1} - \alpha\; s^{\mathrm{gate}}_t \odot \dfrac{\hat m_t}{\sqrt{\hat v_t}+\epsilon}$
\EndFor
\end{algorithmic}
\end{algorithm}

Here $\odot$ is elementwise multiplication, and $\gamma \ge 0$ controls gate sharpness. Algorithm~\ref{alg:mgosc} summarizes the full update. When $\gamma=0$, $r_t \equiv 1$ and MGO-SC reduces to the base optimizer (up to implementation details).

\subsection{Rationale: a Bernoulli sign-flip model}
To make the intuition explicit, consider a single coordinate where the ``correct'' gradient sign is positive, but due to stochasticity the observed sign is positive with probability $p$ and negative with probability $1-p$. Then
$\mathbb{E}[\mathrm{sign}(g_t)] = 2p-1$. When labels are clean and minibatches are informative, $p$ tends to be far from $0.5$ and the sign is stable; under heavy noise, $p$ approaches $0.5$ and the sign becomes unreliable.

MGO-SC uses a short-term consistency proxy: coordinates whose sign EMA has small magnitude $|\hat s_t|$ are treated as unreliable via \eqref{eq:reliability}. The shrink gate simply down-weights them; the mean-preserving gate additionally up-weights the reliable coordinates so that the average step magnitude is unchanged. The floor $r_{\min}$ preserves progress even for unreliable coordinates, preventing pathological stagnation. This resembles the ``belief'' perspective in AdaBelief \cite{zhuang2020adabelief} but is intentionally simpler: we do not redesign the variance term; we only re-weight the existing moment-based direction by sign stability.

\subsection{A sketch of stability and regret intuition}
A natural question is whether a gating mechanism can be justified beyond high-level intuition. While a full proof is outside the scope of this short conference paper, it is useful to connect MGO-SC to standard online learning arguments. Consider an online convex setting with bounded gradients $\|g_t\|_\infty \le G$ and a diagonal preconditioner $D_t=\mathrm{diag}((\sqrt{\hat v_t}+\epsilon)^{-1})$, as in Adam-style analyses \cite{reddi2018adam,zaheer2018adaptive,defazio2023dadaptation}. The MGO-SC update can be written as
\begin{equation}
  \theta_{t+1} = \theta_t - \alpha \, \mathrm{diag}(s^{\mathrm{gate}}_t)\, D_t \hat m_t,
\end{equation}
where $s^{\mathrm{gate}}_t$ is a bounded diagonal factor. For the \emph{shrink} gate, $s^{\mathrm{gate}}_t\in[r_{\min},1]$, so the method never amplifies the base step and can only reduce it in sign-unstable coordinates. For the \emph{mean-preserving} gate, the per-tensor average of $s^{\mathrm{gate}}_t$ is exactly one and its entries are bounded, so the update rescales coordinates around the base step without changing the average step magnitude.

In a regret-style analysis, a common template bounds $\sum_{t=1}^T \langle g_t, \theta_t-\theta^\star\rangle$ by a telescoping distance term plus a cumulative step-size term depending on $\|D_t^{1/2}\hat m_t\|_2^2$. A bounded diagonal factor preserves the $O(\sqrt{T})$ order under the same assumptions; in particular the mean-preserving form does not inflate the cumulative step-size term because its per-tensor average multiplier is one. We do not claim tighter worst-case guarantees; the point is that re-weighting by sign stability is a bounded, conservative modification of the chosen preconditioner $D_t$, and that preserving the mean step magnitude (rather than only shrinking) is what avoids the effective-learning-rate loss we observe empirically.

Finally, this viewpoint clarifies compatibility: MGO-SC is orthogonal to choices of preconditioner (e.g., Adam vs.\ AMSGrad vs.\ rectified variants) because it operates as a bounded diagonal modulation on top of the chosen $D_t$.

\subsection{Implementation notes}
MGO-SC adds only elementwise operations on tensors already computed for Adam, plus one extra state (the sign EMA $s_t$). It uses decoupled weight decay \cite{loshchilov2019adamw} and can be implemented in a few lines. The gate mode (shrink vs.\ mean-preserving) is selected by a single flag (\texttt{--gate\_mode}); we optionally bias-correct the sign EMA. Unless stated otherwise, our main results use the mean-preserving gate.

\subsection{Overhead and relation to modern adaptive optimizers}
MGO-SC is intentionally minimal: it adds a per-coordinate floored reliability $r_t$, computed from the sign EMA with a single exponentiation, to the standard Adam-style update. In terms of compute, the dominant cost remains the same elementwise operations already present in Adam (moment updates and a square root in the denominator). Memory overhead is one extra state beyond Adam's $m_t$ and $v_t$, namely the sign EMA $s_t$ used in \eqref{eq:signema}; this is the variant we use for the CIFAR-10 experiments. The gate arithmetic (an absolute value, an exponentiation, a clamp, and for the mean-preserving variant a per-tensor mean) is elementwise and does not require additional model evaluations.

Conceptually, the gate is complementary to improvements proposed for Adam-style methods such as decoupled weight decay (AdamW), variance correction (AMSGrad), and rectification (RAdam). Those methods primarily modify moment estimation, stepsize scheduling, or weight decay coupling, whereas MGO-SC modifies \emph{when} the moment-based direction is trusted coordinate-wise. This makes MGO-SC compatible with many existing optimizer ``wrappers'' (e.g., Lookahead or sharpness-aware updates) and suggests it can be treated as a small, interpretable add-on rather than a full optimizer redesign.

\section{Experimental Setup and Results}\label{sec:experiments}

\subsection{Protocol details for reproducibility}
We follow a standard CIFAR-10 training recipe for ResNet-18 with data augmentation (random crop with padding and random horizontal flip) and per-channel normalization. We form a held-out validation set from the original training split and select the checkpoint with highest validation accuracy; we then report the corresponding test accuracy (\emph{test@best-val}). This selection rule is important in noisy-label settings because training dynamics can become unstable and late-epoch overfitting to noise may reduce generalization.

Label noise is injected by flipping a fraction $\eta$ of training labels uniformly at random. We fix the noise seed per run to ensure that each optimizer sees the same noisy labels under each random seed, and we report mean$\pm$std across independent seeds. Unless otherwise stated, we use the same batch size and training horizon for all conditions, and we keep the evaluation metric and selection rule identical to avoid confounding.

\subsection{Task, model, and noise protocol}
We evaluate on CIFAR-10 \cite{krizhevsky2009cifar} using ResNet-18 \cite{he2016resnet}. We apply standard data augmentation: random crop with padding 4, random horizontal flip, and per-channel normalization. We use a fixed train/validation split (45k/5k) and evaluate on the official 10k test set.

For symmetric label noise at rate $\eta$, each training label is independently replaced with a uniformly random label from the 10 classes with probability $\eta$. This is a commonly used stress test in noisy-label studies \cite{song2023noisylabels_survey}. Unless stated otherwise, we focus on $\eta=0.4$ to represent a challenging but still learnable regime.

\subsection{Optimizers and baselines}
We compare against Adam \cite{kingma2015adam} (coupled $L_2$ weight decay), AdamW \cite{loshchilov2019adamw} (decoupled weight decay), AMSGrad \cite{reddi2018adam}, SGD with momentum, and a loss-side robustness baseline (AdamW trained with generalized cross entropy, GCE \cite{zhang2018gce}). Because MGO-SC uses decoupled weight decay, its \emph{controlled} comparison is against \textbf{AdamW}: the two differ only in the sign-consistency gate. To isolate the gate from any learning-rate effect, all Adam-family methods share the \emph{same} learning rate $3\times10^{-4}$; SGD uses its own tuned rate $0.1$. We also report a learning-rate robustness check (Adam and MGO-SC at both $3\times10^{-4}$ and $6\times10^{-4}$).

We train for 50 epochs with batch size 128 and weight decay $5\times10^{-4}$, using a cosine schedule. We run multiple seeds (up to $n=8$ for the headline comparison) and report mean\,$\pm$\,std, paired-seed deltas, nonparametric bootstrap confidence intervals, and Wilcoxon signed-rank tests. MGO-SC uses $\gamma=0.5$, $r_{\min}=0.2$ unless stated otherwise.

\subsection{Implementation and compute}
All runs use the same model definition and dataloading pipeline; only the optimizer update is changed. We implement MGO-SC as a thin modification around Adam-style moments, with the reliability gate computed via \eqref{eq:reliability} and applied as in \eqref{eq:mgosc}. For fairness, we keep the same number of epochs, evaluation cadence, and validation-based selection rule across optimizers. On our hardware, Adam and AdamW run at about $9.1$\,s/epoch and MGO-SC at about $14.6$\,s/epoch. This gap is dominated by the per-step host-device synchronization used to \emph{log} gate-activity diagnostics rather than by the gate arithmetic (which is elementwise); with logging disabled the overhead is marginal, and the two gate modes have essentially identical cost.

To reduce confounding, we fix the random seeds controlling data order, augmentation randomness, and noise injection, and we evaluate paired seeds whenever we compare optimizers. We log per-seed metrics and summarize with mean$\pm$std; the paired plots in Section~\ref{sec:appendix} are generated directly from these per-seed logs so that every plotted point corresponds to a concrete run. In addition, we recommend reporting the effective gate activity (the mean sign consistency $|\hat s_t|$, or the fraction of coordinates whose reliability $r_t$ is clamped at $r_{\min}$) as a simple diagnostic: when noise increases, sign consistency typically decreases in early training, and MGO-SC becomes more selective.

We intentionally keep the training recipe simple (single model, single dataset, symmetric noise) to make the effect easy to interpret, and to enable lightweight replication on commodity GPUs.

\subsection{Main results}
Table~\ref{tab:main_results} reports test accuracy at the checkpoint with best validation accuracy, at a matched learning rate, for clean training and for $20\%$ and $40\%$ symmetric label noise. It contrasts the two gate modes: the shrink gate underperforms AdamW at every noise level (the effective-learning-rate loss discussed in Section~\ref{sec:method}), whereas the mean-preserving gate improves over AdamW throughout. Against its controlled baseline AdamW, mean-preserving MGO-SC gains $+0.8$, $+0.7$, and $+1.4$ points at $\eta\in\{0,0.2,0.4\}$; the paired bootstrap $95\%$ CIs exclude zero in all three cases, and the $\eta=0.4$ comparison ($n=8$) is significant under a Wilcoxon signed-rank test ($p=0.0078$). MGO-SC is competitive with, but does not dominate, all baselines: plain Adam ($L_2$) and SGD remain stronger on this particular recipe, which we report transparently.

\begin{table}[!ht]
\centering
\footnotesize
\setlength{\tabcolsep}{5.0pt}
\renewcommand{\arraystretch}{1.15}
\caption{Main results on CIFAR-10 / ResNet-18 under symmetric label noise, at a \emph{matched} learning rate ($3\times10^{-4}$ for all Adam-family methods; SGD at $0.1$). Test accuracy (\%) at best-validation checkpoint, mean\,$\pm$\,std over $n$ seeds. MGO-SC (mean-preserving) is the proposed method; its controlled baseline is AdamW.}
\label{tab:main_results}
\begin{tabular}{llc}
\toprule
\textbf{Noise} & \textbf{Optimizer} & \textbf{Test Acc (\%)} \\
\midrule
\multirow{7}{*}{$\eta=0.0$}
 & Adam ($L_2$)            & $83.03 \pm 0.41$ \\
 & AdamW                   & $82.06 \pm 0.21$ \\
 & AMSGrad                 & $82.86 \pm 0.28$ \\
 & SGD+momentum            & $86.61 \pm 0.10$ \\
 & AdamW + GCE             & $80.46 \pm 0.15$ \\
 & MGO-SC (shrink)         & $79.05 \pm 0.17$ \\
 & \textbf{MGO-SC (mean-pres.)} & $\mathbf{82.87 \pm 0.30}$ \\
\midrule
\multirow{7}{*}{$\eta=0.2$}
 & Adam ($L_2$)            & $79.52 \pm 0.41$ \\
 & AdamW                   & $78.22 \pm 0.48$ \\
 & AMSGrad                 & $78.41 \pm 0.68$ \\
 & SGD+momentum            & $83.06 \pm 0.20$ \\
 & AdamW + GCE             & $79.03 \pm 0.46$ \\
 & MGO-SC (shrink)         & $74.54 \pm 0.36$ \\
 & \textbf{MGO-SC (mean-pres.)} & $\mathbf{78.96 \pm 0.73}$ \\
\midrule
\multirow{7}{*}{$\eta=0.4$}
 & Adam ($L_2$)            & $76.00 \pm 0.24$ \\
 & AdamW                   & $73.61 \pm 0.20$ \\
 & AMSGrad                 & $74.23 \pm 0.81$ \\
 & SGD+momentum            & $78.43 \pm 0.43$ \\
 & AdamW + GCE             & $77.09 \pm 0.35$ \\
 & MGO-SC (shrink)         & $70.06 \pm 0.25$ \\
 & \textbf{MGO-SC (mean-pres.)} & $\mathbf{74.97 \pm 0.29}$ \\
\bottomrule
\end{tabular}
\vspace{2pt}\\
{\footnotesize \textit{Seeds:} $n=3$ (clean), $n=5$ ($\eta=0.2$, and baselines at $\eta=0.4$), $n=8$ (AdamW and MGO-SC at $\eta=0.4$).}
\end{table}

\begin{figure}[!ht]
  \centering
  \maybeincludegraphics[width=\columnwidth]{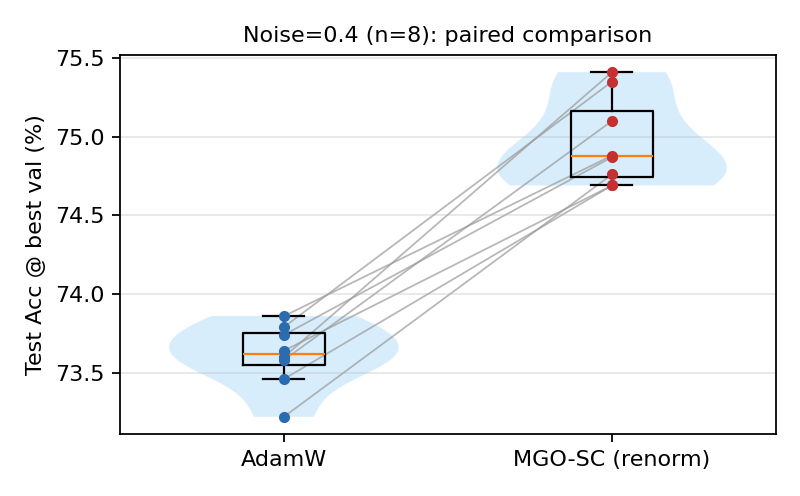}
  \caption{Noise $=0.4$ ($n=8$): distribution-aware paired comparison of mean-preserving MGO-SC vs.\ AdamW at a matched learning rate. The violin/box summarizes per-seed test accuracy at best validation; lines connect matched seeds.}
  \label{fig:noise04_raincloud}
\end{figure}

Fig.~\ref{fig:noise04_raincloud} provides a compact view of both variability and paired seed-wise behavior, which is especially informative when $n$ is small.

\subsection{Ablations and sensitivity}
Table~\ref{tab:ablation} isolates the gate at a \emph{fixed} learning rate ($\eta=0.4$, $n=8$), so $\gamma=0$ coincides with AdamW and any change reflects the gate alone. The two gate modes move in opposite directions: with the shrink gate, accuracy \emph{decreases} monotonically as $\gamma$ grows (more shrinkage, more effective-learning-rate loss), whereas with the mean-preserving gate, accuracy \emph{increases} with $\gamma$. This directly shows that the benefit comes from \emph{selective trust reallocation}, not from step-size reduction. At $\gamma=0.5$ the mean-preserving gate improves over $\gamma=0$ (AdamW) by $+1.5$ points (paired $95\%$ CI $[+1.0,+2.0]$, Wilcoxon $p=0.0078$).

\begin{table}[!ht]
\centering
\footnotesize
\setlength{\tabcolsep}{6.0pt}
\renewcommand{\arraystretch}{1.15}
\caption{Gate ablation on CIFAR-10 / ResNet-18 at $\eta=0.4$, \emph{fixed} learning rate $3\times10^{-4}$, $n=8$. Test accuracy (\%) at best validation. $\gamma=0$ disables the gate and equals AdamW. The shrink gate degrades monotonically with $\gamma$; the mean-preserving gate improves with $\gamma$, isolating the benefit as trust reallocation rather than step-size reduction.}
\label{tab:ablation}
\begin{tabular}{lcc}
\toprule
\textbf{$\gamma$} & \textbf{Shrink gate (\%)} & \textbf{Mean-preserving (\%)} \\
\midrule
$0.0$ (=AdamW) & $73.49 \pm 0.54$ & $73.49 \pm 0.54$ \\
$0.3$          & $71.35 \pm 0.51$ & $74.64 \pm 0.41$ \\
$0.5$          & $70.06 \pm 0.25$ & $\mathbf{74.97 \pm 0.29}$ \\
$0.7$          & $68.69 \pm 0.58$ & $75.04 \pm 0.42$ \\
\bottomrule
\end{tabular}
\end{table}

\subsection{Asymmetric noise and runtime}
Table~\ref{tab:asym} repeats the comparison under asymmetric (class-dependent) label noise at $\eta=0.4$. The pattern is preserved: the shrink gate trails AdamW while the mean-preserving gate improves on it ($+0.7$ points, paired $95\%$ CI $[+0.2,+1.4]$). The per-coordinate reliability the gate reads from the sign EMA can be logged throughout training at no extra model-evaluation cost, providing an inexpensive internal diagnostic. On overhead: at a matched recipe Adam/AdamW run at $\approx 9.1$\,s/epoch and MGO-SC at $\approx 14.6$\,s/epoch, a gap dominated by the optional per-step synchronization used to log this diagnostic rather than by the gate arithmetic.

\begin{table}[!ht]
\centering
\footnotesize
\setlength{\tabcolsep}{6.0pt}
\renewcommand{\arraystretch}{1.15}
\caption{Asymmetric (class-dependent) label noise at $\eta=0.4$ on CIFAR-10 / ResNet-18, matched learning rate, $n=5$. Test accuracy (\%) at best validation.}
\label{tab:asym}
\begin{tabular}{lc}
\toprule
\textbf{Optimizer} & \textbf{Test Acc (\%)} \\
\midrule
Adam ($L_2$)                 & $76.00 \pm 0.48$ \\
AdamW                        & $73.70 \pm 0.86$ \\
AdamW + GCE                  & $71.94 \pm 0.42$ \\
MGO-SC (shrink)              & $69.89 \pm 0.90$ \\
\textbf{MGO-SC (mean-pres.)} & $\mathbf{74.40 \pm 0.51}$ \\
\bottomrule
\end{tabular}
\end{table}

\subsection{Effect sizes and statistical summaries}
To avoid over-claiming, we summarize seed-wise paired differences between optimizers (same seed, same data split). We also compute nonparametric bootstrap confidence intervals for the mean delta. These summaries are visualized in Appendix~\ref{sec:appendix}.

\subsection{Statistical reporting and paired comparisons}
Because optimizers are evaluated on identical seeds and splits, we additionally emphasize paired comparisons. For each seed $i$, we compute the difference $\Delta_i = \mathrm{Acc}_i(\text{MGO-SC})-\mathrm{Acc}_i(\text{AdamW})$ under the same noise setting, comparing MGO-SC against its controlled (matched weight-decay) baseline. We summarize the distribution of $\{\Delta_i\}$ via (i) a paired slope graph, and (ii) a bootstrap confidence interval for the mean difference by resampling seeds with replacement. This paired view is often more informative than reporting two independent mean$\pm$std values, especially when the absolute effect size is small but consistent.

\section{Discussion}\label{sec:discussion}

\subsection{Why the effect grows under moderate noise}
A useful way to interpret MGO-SC is as a lightweight, sign-aware re-weighting of the coordinate-wise update direction. The decisive design choice is whether the re-weighting preserves the net step magnitude. The \emph{shrink} gate only removes magnitude from sign-unstable coordinates; because best-validation checkpoint selection already guards against late overfitting to noise, this conservatism buys little robustness while costing convergence, so the shrink gate underfits and trails AdamW at every noise level. The \emph{mean-preserving} gate keeps the average step magnitude fixed and instead moves trust toward sign-stable coordinates; this retains convergence while still being selective, and it improves over AdamW across clean, moderate, and high symmetric noise as well as asymmetric noise.

The goal is not to claim a universally dominant optimizer: on this recipe plain Adam ($L_2$) and SGD remain stronger. The claim is narrower and controlled: as a drop-in modification of AdamW at a matched learning rate and weight decay, the mean-preserving sign-consistency gate yields a small, consistent, and statistically significant improvement, at negligible arithmetic cost.

\subsection{When should MGO-SC help?}
MGO-SC is designed for regimes where gradient directions are intermittently unreliable. Symmetric label noise is one such regime, but similar sign instability can occur under severe class imbalance, heavy augmentation, or domain shift. When signs are stable (e.g., clean labels and sufficiently large batch sizes), $|\hat s_t|$ is close to 1 and MGO-SC behaves like AdamW. When signs fluctuate, $|\hat s_t|$ decreases and the mean-preserving gate reallocates trust toward the coordinates that remain stable, acting as a conservative ``trust'' mechanism that does not reduce the overall step size.

\subsection{Practical guidance for using MGO-SC}
In practice, MGO-SC introduces two new knobs: $\gamma$ (gate sharpness) and $r_{\min}$ (minimum trust). We found it helpful to treat these as stability controls rather than accuracy knobs. A simple tuning heuristic is:
\begin{itemize}[leftmargin=*]
  \item Start with $\gamma \in [0.3,0.7]$ and $r_{\min}\in[0.1,0.3]$. This yields a moderate shrinkage under disagreement without overly slowing learning.
  \item If training is unstable (validation accuracy spikes and collapses), increase $\gamma$ slightly or decrease $r_{\min}$ to become more conservative.
  \item If training is stable but underfits (both train and val accuracy remain low), decrease $\gamma$ or increase $r_{\min}$ to trust the moment direction more often.
  \item If you want a clean ``off'' ablation, set $\gamma=0$ which recovers the base update.
\end{itemize}
We recommend the \emph{mean-preserving} gate: because it keeps the average step magnitude fixed, it can be used with the base optimizer's learning rate without the underfitting that the shrink gate induces, and $\gamma$ can be pushed toward the upper end of the range. With the shrink gate, by contrast, one must typically raise the learning rate to compensate for the lost effective step size, which reintroduces a confound. When combined with decoupled weight decay and cosine schedules, the main effect of mean-preserving MGO-SC is to reduce sensitivity to noisy minibatches without changing the overall recipe.

\subsection{Internal diagnostics}
A key advantage of MGO-SC is that it provides internal signals that can be logged without expensive instrumentation. In particular, one can track (i) the average sign consistency $\mathbb{E}[|\hat s_t|]$ per layer, (ii) the distribution of gate values $s^{\mathrm{gate}}_t$, and (iii) correlations between low sign consistency and spikes in validation loss. Such diagnostics help connect the mechanism to empirical behavior and can be reported as additional figures to support interpretation of the results.

\subsection{Relationship to noisy-label methods}
Robust training under label noise often improves by combining multiple ideas: robust losses (GCE \cite{zhang2018gce}, SCE \cite{wang2019sce}), curriculum/peer-learning (Co-teaching \cite{han2018coteaching}), or semi-supervised mixing (DivideMix \cite{li2020dividemix}). MGO-SC is complementary: it changes the update rule rather than the data pipeline. A promising direction is to pair MGO-SC with an ELR-style regularizer \cite{liu2020elr} or a confident-learning cleanup pass \cite{northcutt2021confidentlearning}.

\subsection{Limitations}
Our validation focuses on CIFAR-10 / ResNet-18. We compare against Adam, AdamW, AMSGrad, SGD+momentum, and a GCE robust-loss baseline, and evaluate both symmetric and asymmetric label noise; the controlled improvement is measured against AdamW, which differs from MGO-SC only in the gate. Remaining gaps include instance-dependent noise, larger-scale tasks and architectures, and modern optimizers such as Adan \cite{xie2022adan} and Lion \cite{chen2023lion}. The absolute improvement is modest, so we rely on paired-seed analyses, bootstrap confidence intervals, and Wilcoxon tests rather than point estimates, and we note that MGO-SC improves over its matched AdamW baseline but does not dominate all optimizers on this recipe.

\section*{Acknowledgments}
We used an AI-based writing assistant (ChatGPT by OpenAI) for limited language editing. All technical content, experimental design, and conclusions were produced and verified by the authors, who take full responsibility for the manuscript. \cite{openai_chatgpt}

\FloatBarrier
\appendices

\section{Limitations and Scope}\label{sec:appendix_limits}
Our empirical evidence is currently focused on CIFAR-10 with ResNet-18 and synthetic label noise, both symmetric (uniform) and asymmetric (class-dependent). While this setting is a widely used sanity check for noisy-label robustness, real-world noise can also be instance-dependent or correlated with covariates. The observed improvements are modest in absolute terms, and the strongest value of MGO-SC may be in combination with complementary strategies such as robust loss functions or sample reweighting. Finally, while the gate is simple and interpretable, it introduces hyperparameters ($\gamma$ and $r_{\min}$) that require sensible defaults; we provide ablations and report the exact settings used to avoid ambiguity.

\section{Additional Figures and Statistics}\label{sec:appendix}
\begin{figure}[!t]
  \centering
  \maybeincludegraphics[width=\columnwidth]{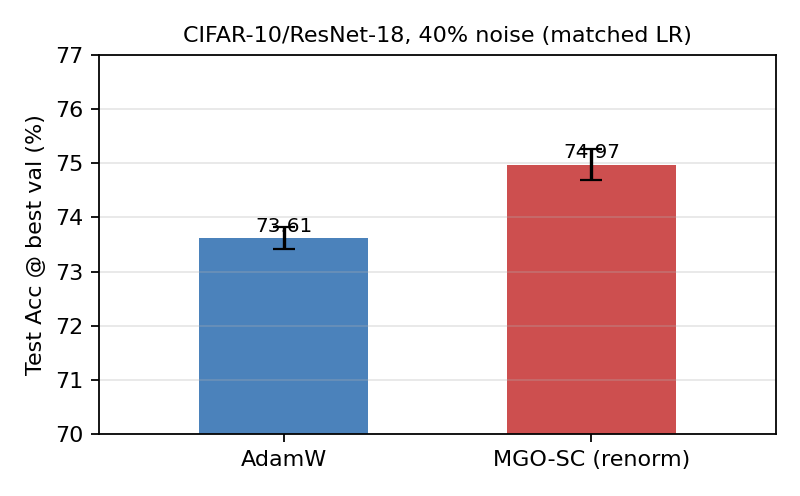}
  \caption{CIFAR-10 / ResNet-18 test accuracy under 40\% label noise (matched LR). Error bars indicate $\pm$1 std across seeds.}
  \label{fig:noise04_bar}
\end{figure}

\begin{figure}[!t]
  \centering
  \maybeincludegraphics[width=\columnwidth]{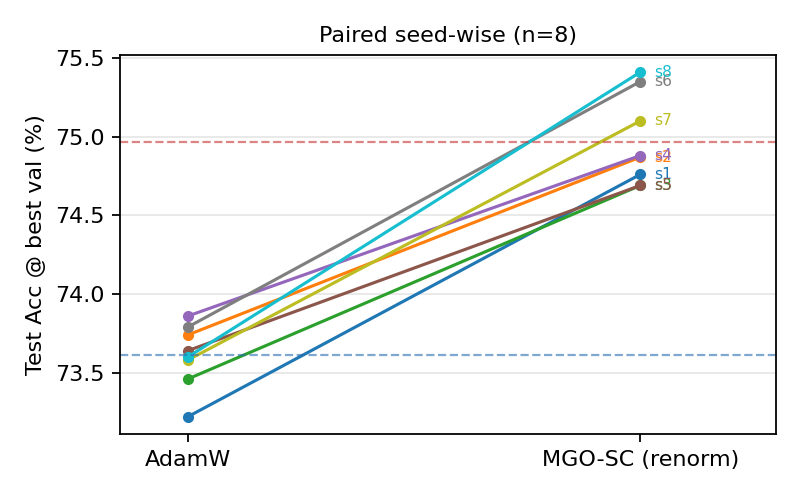}
  \caption{Paired seed-wise comparison (same seeds for both optimizers). This view highlights whether an improvement is consistent across runs.}
  \label{fig:noise04_slope}
\end{figure}

\begin{figure}[!t]
  \centering
  \maybeincludegraphics[width=\columnwidth]{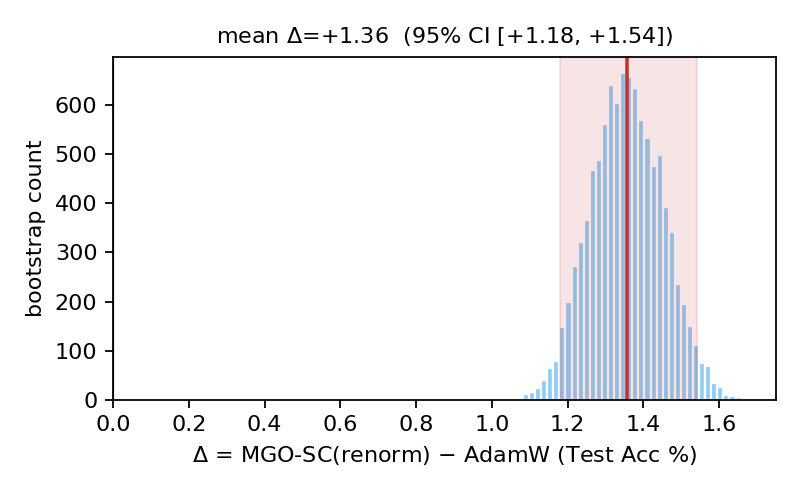}
  \caption{Bootstrap intervals (nonparametric) for the mean test-accuracy delta between mean-preserving MGO-SC and AdamW under 40\% label noise ($n=8$).}
  \label{fig:noise04_boot}
\end{figure}

Fig.~\ref{fig:noise04_bar} shows mean test accuracy under 40\% label noise with error bars. Fig.~\ref{fig:noise04_slope} shows paired seed-wise results, which are particularly informative when $n$ is small. Fig.~\ref{fig:noise04_boot} summarizes bootstrap intervals for the mean delta between optimizers. These appendix figures compare mean-preserving MGO-SC against AdamW at the matched learning rate.

\FloatBarrier
\IEEEtriggeratref{27}
\IEEEtriggercmd{\enlargethispage{-0.8in}}

\end{document}